# Emerging Artificial Intelligence Applications in Spatial Transcriptomics Analysis

Yijun Li[1], Stefan Stanojevic[2], Lana X. Garmire[1,2] *

[1]Department of Biostatistics, University of Michigan, Ann Arbor, MI, USA

[2]Department of Computational Medicine and Bioinformatics, University of Michigan, Ann Arbor, MI, USA

*Corresponding author, email: lgarmire@med.umich.edu

## Abstract

Spatial transcriptomics (ST) has advanced significantly in the last few years. Such advancement comes with the urgent need for novel computational methods to handle the unique challenges of ST data analysis. Many artificial intelligence (AI) methods have been developed to utilize various machine learning and deep learning techniques for computational ST analysis. This review provides a comprehensive and up-to-date survey of current AI methods for ST analysis.

**Keywords:** spatial transcriptomics; artificial intelligence; deep learning; machine learning

## Introduction

ST refers to transcriptome technologies that can preserve the spatial context and gene expression profiles of the tissue sample. The past years have witnessed tremendous growth in the field of ST (Figure 1(a)). Depending on the data generation method, ST technologies can be divided into NGS-based (next-generation sequencing) and image-based approaches [1]. NGS-based ST technologies obtain spatially-resolved data by attaching spatial barcodes with fixed locations to tissue sections. As a result, each spot captured by NGS-based ST datasets usually contains multiple cells. Many NGS-based ST methods have been developed, including Visium by 10X Genomics [2], GeoMx by NanoString [3], Slide-Seq [4], etc. Image-based methods obtain RNA transcripts via either *in-situ* sequencing or *in-situ* hybridization and retain the spatial information of the cells through images of the stained tissue sample. Image-based ST techniques such as STARMap [5], merFISH [6], and seqFISH+ [7] often achieve single-cell or subcellular resolution. Typically, an ST dataset consists of a gene expression matrix where each row represents a gene and each column a spot/cell, and a spatial location matrix where the spatial coordinates of the spots/cells are recorded (Figure 1(b)).

Many new computational challenges for ST analysis come along with the new ST technologies. Since the spatial context of tissues is highly relevant to gene expression, cell type distribution, cell-cell communication, and cell function, there is a need for novel computational methods that can analyze ST data while taking full advantage of the added spatial information. In recent years, machine learning and deep learning methods have become increasingly popular in single-cell transcriptomics analysis due to their ability to analyze large data using sophisticated model architectures. In this article, we review the many deep learning and machine learning methods that have been developed to tackle different aspects of ST analysis, including detecting spatially variable genes, clustering, communication analysis, deconvolution, and enhancement (Table 1 and Figure 1(b)). Specifically, we focus on methods that directly work with ST data. Computational tools that infer spatial location or spatial gene expression based on other data types were excluded. For more general surveys on ST, readers are encouraged to refer to the work of Rao et. al. [1], Lu et. al. [8], Atta. et. al. [9], etc.

## AI Methods for Spatially Variable Gene Detection

Detecting spatially variable genes (SVGs) is an essential step of ST analysis. SVGs are defined as genes whose expression patterns across physical space are significantly distinct. SVGs can be novel markers for specific cell types; they can also be used to refine expression histology and further elucidate the spatial architecture of the data. Most SVG detection methods are hypothesis testing frameworks based on either spatial point process models [10] or Gaussian Processes [11–13]. However, there have also been some machine-learning-based approaches developed for detecting SVGs. Such methods utilize machine learning techniques to improve the statistical framework by compressing the data and reducing computational burden [14], or adapt SVG detection to a binary computer vision problem [15].

**SOMDE** [14] is a hybrid machine learning and statistical method to detect SVG based on self-organizing maps (SOM) and the Gaussian Process model. The SOM clusters neighboring spatial spots and outputs condensed spatial nodes while preserving the original topological structure and relative spot densities. The meta-gene expression of the compressed nodes is computed as the weighted average of the maximum and the average expression values of the cluster of spots corresponding to each node. The compressed ST data are then fit to a Gaussian Process model similar to spatialDE [11]. Given the spatial coordinates of the compressed SOM nodes $\tilde{X}$, the meta expression of a gene on the SOM scale $\tilde{y}$ is modeled using Gaussian Process (see Eq. (1)). The kernel function is decomposed as the sum of a squared exponential kernel of the spatial locations ($\Sigma_{k(\tilde{X},\tilde{X}'|\theta)}$) and random noise ($\delta \cdot I$). Similar to spatialDE [11], SOMDE constructs a null model under which the spatial variation of the gene is random (see Eq. (2)). The significance of each gene's spatial variation is determined using a likelihood ratio test. The nominal p-value of each gene is adjusted for multiple testing. Compared to other statistical methods, SOMDE is 5-50 times more efficient. Its first step enables data compression, which lessens the computational burden of the subsequent Gaussian Process model without losing crucial spatial structures.

$$\text{Full model: } P(\tilde{y} \mid \tilde{X}, \theta) = N(\tilde{y} \mid \mu \cdot 1, \sigma_S^2 \cdot \Sigma_{k(\tilde{X},\tilde{X}'|\theta)} + \delta \cdot I) \quad (1)$$

$$\text{Null model: } P(\tilde{y} \mid \tilde{X}, \theta) = N(\tilde{y} \mid \mu \cdot 1, \delta \cdot I) \quad (2)$$

**scGCO** [15] identifies SVG by optimizing Markov Random Fields with graph cut. scGCO treats SVG detection as an image segmentation problem. For each gene, scGCO builds a graph representation of the spatial information using Delaunay Tessellation [16]. This graph representation naturally induces an underlying Markov Random Field model (MRF). The MRF is clustered into two subgraphs based using max-flow min-cut algorithm. The statistical significance of the identified spatial expression pattern is determined using a homogeneous spatial Poisson distribution. scGCO can scale up in dimensionality to handle three-dimensional ST data such as STARMaps [5]. In addition, it does not assume prior assumptions on data distribution and is theoretically guaranteed to find the global optimal solution. When applied to the Mouse Olfactory Bulb dataset from Stahl et. al. [17], scGCO detected significantly more SVGs than spatialDE [11] while using less computational memory.

## AI Methods for Clustering Analysis of Spatial Transcriptomics Data

Clustering analysis is an integral step in transcriptome data analysis. In the context of ST data, clustering spots or genes involves grouping together spots or genes with similar transcriptional profiles and spatial information profiles. Clustering is important for annotating cell types, understanding tissue structure, identifying co-expressed gene modules and many downstream analyses such as contextualizing trajectory inference and cell-cell communication. To this end, many deep learning methods leveraging convolutional neural networks (Figure 2(b)), graph convolutional neural networks (Figure 2(c)), autoencoders (Figure 2(d)) and their variants such as variational autoencoders (Figure 2(d)) have been developed to cluster ST data [18–24].

**SEDR** [21] is an unsupervised autoencoder model for extracting low-dimensional latent embeddings of ST data. SEDR has two components. First, a deep autoencoder learns the latent representation of gene expression. Then SEDR constructs a spatial graph based on the Euclidean distances between the spots/cells and represents the graph via a binary adjacency matrix. A variational graph autoencoder combines the constructed spatial graph and the latent embedding from the deep autoencoder model and learns the latent representations of spatial information. The latent gene and spatial embeddings are then concatenated and further fed through an iterative deep clustering algorithm [25]. The resulting joint embedding can then be used to perform clustering analysis, which was shown to have increased accuracy compared to further downstream analysis such as Seurat [26], Giotto [27], stLearn [18], and Bayespace [28]. SEDR can also be applied for trajectory analysis, batch correction, and visualization.

**CoSTA** [20] is an unsupervised gene clustering method that learns spatial similarity between genes using convolutional neural network (CNN). The CoSTA workflow is inspired by DeepCluster, which jointly learns the neural network parameters with the clustering labels. In the CoSTA framework, the expression of each gene is represented as a matrix whose rows and columns indicate the spatial coordinates of the spots. The gene expression matrices are forwarded through a neural network with three convolutional layers, each followed by a batch normalization layer and a max pooling layer. The corresponding matrix output for each gene is then flattened into a vector. Such vectors can be interpreted as a spatial representation of the corresponding gene. The combined spatial representation vectors are then normalized using

L2-normalization, dimension reduced using UMAP, and clustered using Gaussian Mixture Modeling (GMM). The final spatial representation vectors learned by the CNN can be used for downstream analyses such as gene clustering, co-expression analysis, SVG identification, visualization, etc. When studying gene-gene relationships, CoSTA emphasizes general spatial patterns in learning representations of each gene, enabling more biologically meaningful results than simply focusing on the exact overlap of cells. The authors showed that CoSTA tended to provide more specific results than other spatial gene analysis methods such as spatialDE [11] and SPARK [12], suggesting that CoSTA has advantages in cases where users would like to narrow down selected genes for further analysis. Since CoSTA is not dependent on the strict overlap of spots, it can also be helpful in cases where gene matrices are not based on exactly the same tissue but neighboring samples.

**STGATE** [23] is a graph attention autoencoder model that clusters the spots/cells in ST data and detects spatial domains. STGATE constructs a binary spatial neighbor network (SNN) based on the pairwise spatial distances between spots. The SNN has the flexibility to be cell-type-aware by pruning the network with pre-clustered gene expression. The gene expression profile and the spatial neighborhood network are then fed into a graph attention autoencoder. The encoder learns a low-dimensional embedding of the gene expression profile spatial information. The graph attention mechanism allows the model to estimate edge weights and update the SNN adaptively. As a result, the authors showed that STGATE improved the accuracy of spatial domain identification and mitigating technical noise in ST data.

**RESEPT** [22] is a deep learning framework that reveals tissue architecture by clustering ST data. RESEPT can take either gene expression information or RNA velocity as input. A spatial graph is built based on pairwise spot distance and gene expression. The Euclidean distance between neighboring spots are represented as edge weights, and the gene expression at each spot are represented as node attributes. Such a graph is then forwarded through a graph autoencoder; the encoder portion embeds the graph into a three-dimensional representation using two graph convolution layers; the decoder reconstructs the graph through a sigmoid activation of the inner product of the graph embedding. The three-dimensional output of the encoder is then mapped to an RGB (red, green, blue) image, which naturally induces a visual representation of the spatial gene expression. The image is segmented via a deep convolutional neural network model, consisting of backbone, encoder, and decoder portions. The backbone portion utilizes ResNet101 [29], a deep neural network model, to extract image features; the encoder portion selects multi-scale semantic features from the features generated by ResNet101; finally, the decoder portion aligns the multi-scale semantic features by size and outputs a segmentation map which clusters the spots and reveals tissue architecture. RESEPT allows for direct visualization of spatial expression. The authors showed that RESEPT accurately inferred spatial architecture with visualization. Furthermore, RESEPT can perform spatial-temporal analysis of ST data via RNA velocity analysis.

**spaGCN** [19] is a spatial domain detection method that can integrate histology information with ST data using graph convolutional neural network (GCN). spaGCN integrates the spatial information from ST data and histology information by concatenating the histology pixel values to the spatial coordinate values. The integrated spatial information matrix is then represented as a weighted undirected graph. Each edge weight is identified by applying a Gaussian kernel to the Euclidean distance between the corresponding spots. The gene expression matrix is dimensionally reduced using PCA. spaGCN combines the spatial and gene expression information using a graph convolution layer. The graph convolution layer allows for integration of gene expression information and spatial information while acknowledging the spatial neighborhood structure. The resulting spot representations are then used for iterative clustering to define coherent spatial domains with respect to genetic, spatial and histological information. spaGCN also allows for detecting SVGs or meta-genes by doing differential gene expression analysis between spots in arbitrary target domains and neighboring domains. The authors demonstrated that spaGCN could define spatial domains with coherent gene expression and histology patterns. Furthermore, the domains identified by spaGCN could detect SVGs or meta genes with much clearer spatial expression patterns compared to other SVG detection methods such as spatialDE [11] and SPARK [12].

**stLearn** [18] is a ST analysis pipeline that can cluster the cells/spots, perform spatial trajectory inference, spot-spot interaction analysis, and microenvironment detection. stLearn utilizes Spatial Morphological gene Expression normalization (SME), a deep-learning-based method for normalization, which considers the spatial neighborhood information and morphological structure of the data. SME normalization requires both ST data and H&E images of the tissue as the input. SME normalization operates under the assumption that cells sharing morphological similarities also have more similar transcriptional profiles. The neighborhood of a spot is determined through a disk-smoothing approach. All spots whose center-to-center physical distances to the target spot within an arbitrary length $r$ are considered its neighbors. SME normalization utilizes morphology information by inputting H&E images to a pre-trained ResNet50 [29] network, a very popular deep convolutional neural network for image classification. The pre-trained ResNet50 model extracts a morphological feature vector for each spot. SME normalization then computes the pairwise morphological similarity of spots by taking the cosine distance of their corresponding feature vectors. Finally, the normalized gene expression of a spot is computed as the average of gene expression in each neighboring spot weighted by the morphological similarity score. After SME normalization, stLearn employs a novel two-step clustering technique SMEclust. First, the normalized gene expression data is clustered using standard Louvain clustering [30]. Then, SMEclust applies a two-dimensional k-d tree neighbor search based on the spatial coordinates, dividing broad clusters that span over spatially disjoint areas into smaller sub-clusters. stLearn pipeline further uses the SMEclust results for downstream analysis such as spatial trajectory inference and spot-spot interaction analysis.

**SpaCell** [24] integrates ST with imaging data to make predictions about cell types and disease stages. There are two main models in SpaCell: a representation learning model that describes each spot using both the image information and the gene expression data and a classification model that predicts the disease stage using the two data modalities. Like stLearn, spaCell's representation learning model starts by using a pre-trained ResNet50 CNN model [31] to extract image-based features describing each spot. Then, two different autoencoders are used to reduce the image-based features and the gene expression values to a latent space of the same dimension. Such representations are then concatenated to produce a joint representation vector for each spot, and clustering is performed on such joint representations to distinguish between cell types in an unsupervised manner. Similarly, the classification model applies a pre-trained CNN model to the imaging data and combines this information with gene expression by using a neural network to arrive at disease-stage predictions. It allows for the pre-trained CNN network to be fine-tuned through the training process to better capture biological data's intricacies. Their model is applied to analyze ST data of prostate cancer and amyotrophic lateral sclerosis patients.

## AI methods for Communication Analysis of Spatial Transcriptomics Data

The study of cell-cell or spot-spot communication is essential for studying cellular states and functions. It is well established that communication between cells/spots can be inferred based on gene expression [32–34]. However, the physical location of cells also restricts communications between cells. Several AI methods based on ensemble learning, graph convolutional neural networks (Figure 2(c)) and variational autoencoders (Figure 2(d)) have been developed for communication analysis of ST data, utilizing the added spatial context [35–37].

**GCNG** [35] is a supervised graph convolutional neural network model for inferring gene interactions in cell-cell communication for single-cell ST data. GCNG takes two inputs: the gene expression matrix of a gene pair, and a matrix which encodes the spatial graph based on the ST data. GCNG first computes the pairwise Euclidean distances between all cell pairs to build the spatial graph. A threshold distance value is used to select neighbors. The resulting binary adjacency matrix is then used to calculate a normalized Laplacian matrix, representing the spatial graph input for the GCNG model. The GCNG model is a five-layer graph convolutional neural network, consisting of two graph convolutional layers, a flatten layer, a dense layer and a final classification layer which determines whether the gene pair interacts. The first graph convolutional layer integrates the gene expression and spatial graph and learns embedding features for each cell. The second convolutional layer combines the embedded features of each cell with its neighbors, allowing users to learn indirect graph relationships. GCNG is trained in a supervised approach, using a curated list of interacting ligands and receptors as the ground truth. The authors showed that GCNG could successfully identify known ligand-receptor pairs with much higher accuracy than single-cell Pearson correlation, spatial Pearson correlation, and Giotto [27]. GCNG can be further utilized downstream for functional gene assignment, causal interaction inference, and co-expression analysis.

**NCEM** [36] is a deep generative method that models cell/spot communication in tissue niches. Given a cell, a niche is defined as the cells within an arbitrary radius from the cell's center. NCEM builds a spatial graph based on the Euclidean distance between cells. The NCEM framework takes three inputs: a matrix specifying the expression of each gene in each cell, a matrix specifying observed cell types of all cells, and a matrix specifying batch assignments. NCEM then feeds the input into an autoencoder. The encoder compresses cell-type labels, graph-level predictors and local graph embedding based on the spatial graph to a latent state. The latent state is then reconstructed through a decoder. Depending on the spatial complexity of the data, NCEM accommodates three levels of model complexities: (1) the local graph embedding can be computed through simple indicator embedding functions, which simplifies the model to a generalized linear model that measures linear expression effects of the cell communication; (2)

the local graph embedding is computed through a graph convolutional neural network, making the framework a nonlinear autoencoder that can model non-linear cell interaction; (3) the nonlinear autoencoder can be further extended to a generative variational autoencoder model, which imposes a probability distribution over the latent space and learns the reconstructed data through a likelihood function. This type of model is also capable of modeling latent confounders. Through this flexible framework, NCEM reconciles variance attribution and communication modeling. Currently, NCEM is only applied to assays with subcellular resolution, namely merFISH, immunohistochemistry and imaging mass cytometry.

**MISTy** [37] is a flexible ensemble machine learning method for scalable cell-cell communication analysis. MISTy consists of multiple "views", each representing a different model under a different spatial context. For example, "intraview" is the baseline view that models intracellular gene interactions, "juxstaview" focuses on capturing local cellular niche, and "paraview" captures the effect of tissue structure. The multiple views form a meta model, where the expression of a gene is modeled as the weighted sum of the output of each view. MISTy used random forests [38] as the machine learning model for each view, but the MISTy framework is also flexible to accommodate other algorithms, as long as the algorithm in question is interpretable and can make up ensemble models. Each view is trained independently first; then the meta-model is trained by linear regression. The flexible framework of MISTy allows users to simultaneously study cell-cell communication under different contexts, analyze each view's contribution to the prediction of gene expression, and rank feature importance.

# AI Methods for Deconvolution of Spatial Transcriptomics Data

Depending on the specific ST technology, the generated ST data do not always have single-cell resolution. In addition, since cell type distribution is correlated with their spatial locations, computing cell-type proportions in each spot utilizing both spatial and genomic information is of great interest. Many deep learning methods have been developed for such purposes, either in combination with high resolution H&E images [39]or by integrating scRNA-Seq data [40,41]. The methods utilize diverse methodologies including neural networks (Figure 2(a)), adversarial mechanisms and variational autoencoders (Figure 2(d)).

**Tangram** [39] aligns ST data with scRNA-seq data from the same tissue by learning a soft mapping between the cells assayed by scRNA-seq and the spots in the ST assays. This mapping is learned by optimizing an objective function characterizing the quality of the cell-spot assignments. It considers the difference between spatial cell densities as measured by the ST assay and as predicted by the cell-spot assignments, and it aims to maximize the cosine similarity between the predicted and observed ST measurements. Once the cell-spot assignments are learned, the lower-resolution ST measurements can be deconvolved to infer the cell type composition of each spot and the spatial structure of single-cell datasets can be inferred. This package also provides functionality for incorporating the histological images in the analysis and was even able to visualize the chromatin accessibility information in space by analyzing SHARE-seq [42] data containing matched RNA and chromatin accessibility information from single cells.

**DSTG** [43] is a semi-supervised method for deconvolving ST data. DSTG uses a graph convolutional neural network model. DSTG uses scRNA-Seq data and ST data as input. First, DSTG generates pseudo-ST data by combining the expression of single cells in the scRNA-Seq data. Then, DSTG creates a soft mapping between the pseudo-ST and real ST data. DSTG reduces the dimension of both datasets using canonical correlation analysis. Then, the dimension reduced datasets are used to build a link graph using the mutual nearest neighbors algorithm, capturing the inherent topological structure of the mapping of spots. Finally, DSTG feeds the link graph and concatenation of the pseudo-ST dataset and the real ST dataset into a graph convolutional neural network with multiple convolution layers, effectively learning a latent embedding of the gene expression and local graph structures. The output layer of the graph convolutional neural network predicts both the cell composition of the pseudo and real ST data. The graph convolutional neural network is trained by minimizing the cross-entropy between the two sets of predicted composition. DSTG is an accurate and efficient method. DSTG consistently outperformed SPOTlight [44] in both synthetic and real datasets when benchmarked.

**DestVI** [40] is a Bayesian deep generative model for deconvolution and continuous estimation of cell states of ST data. DestVI consists of two latent variable models (LVMs): one for the reference scRNA-Seq data (scLVM), and the other for the ST data (stLVM). scLVM is quite similar to scVI [45], and it models the gene expression of each gene per cell as a negative binomial distribution. The cell type of each cell and an underlying latent vector describing its variability within each cell type are mapped to the negative binomial model via a neural network.

scLVM learns the distribution for each cell, quantifying the probability of potential cell states. The rate parameter of the distribution is dependent on latent variables that respectively capture technical and biological variations over all possible cell types. Correspondence between the two LVMs is established by sharing the same decoder. DestVI estimates the cell type proportion in each spot and approximates the average cell-state for every cell type in that spot.

**CellDART** [41] is a supervised neural-network-based model for estimating the cell-type composition of spots in non-single-cell resolution ST data. It utilizes both ST data and scRNA-Seq data as the reference and deconvolutes ST data by adapting an ADDA (Adversarial Discriminative Domain Adaptation algorithm) [46], a domain adaptation algorithm that utilizes GAN (Generative Adversarial Network) loss. Cells in scRNA-Seq data are randomly selected to form a coarsened "pseudospot" whose cell-type composition is known. CellDART employs a feature embedder to compute lower-dimensional latent features of ST or reference scRNA-Seq data. The feature embedder is attached to a source classifier model that predicts each spot's cell type composition and a domain classifier that separates the "pseudospot"s from the real ST spots. This domain adaptation mechanism allows CellDART to learn the cell composition in ST data. For the loss function, CellDART uses a loss function based on Kullback-Leibler divergence ($L_S$) and two separate adversials loss functions ($L_{adv,1}$ and $L_{adv,2}$). The feature embedder and the source classifier are first pre-trained using $L_S$. Then the entire CellDART model is trained by iteratively minimizing $L_S$, $L_{adv,1}$ and $L_S$, $L_{adv,2}$. When applied to the Human Dorsolateral Prefrontal Cortex dataset (Visium) [47], CellDART was able to achieve higher AUC (area under curve) values than other deconvolution tools such as Scanorama [48], Cell2location [49], RCTD [50], SPOTlight [44], Seurat [51] and SPOTlight [44].

## AI Methods for Enhancement & Imputation of Spatial Transcriptomics Data

Besides deconvolution, enhancing the spatial gene expression of non-single-cell ST data is another important aspect of computational ST analysis. Such tasks usually require reference data such as high-resolution histological images or single-cell RNA-Seq data. Many deep learning techniques, including fully connected neural networks (see Figure 2(a)), convolutional neural networks (see Figure 2(b)) and autoencoders (see Figure 2(d)) have been developed to enhance the resolution of ST data. We focus on AI methods that use ST data as input. Methods that infer ST data using purely other data types will not be discussed in this section [52].

**XFuse** [53] uses a Bayesian deep generative model to enhance the resolution and impute spatial gene expression with histological images. XFuse assumes the gene expression and histological image share an underlying latent state. The conditional distribution of the gene expression and the histological image given the latent state are negative binomial and Gaussian, respectively. The parameters of these conditional distributions are mapped from the latent state through a neural generator network. XFuse utilizes variational inference to approximate the joint posterior distribution. The underlying tractable distribution parameters are encoded by a convolutional recognition network. The generator and recognition networks form an U-Net-like structure [54].

The latent tissue state is modeled over multiple resolutions to efficiently capture the spatial gene expression of the tissue. XFuse can enhance the resolution of spatial gene expression up to the resolution of the integrated histological image and impute spatial gene expression at missing spots.

**DeepSpaCE** [55] is a semi-supervised learning method that imputes spatial gene expression from H&E images and enhances the resolution of ST data using convolutional neural networks. H&E images are split into sub-images of each spatial spot. Pairs of spot images are forwarded through a deep convolutional neural network with sixteen weight layers, adapted from the VGG16 architecture [56], a very deep convolutional neural network model for image recognition. The output of the VGG16 network predicts either the gene expression or the gene cluster type of the corresponding spot. The authors showed that DeepSpaCE could predict gene expression on missing spatial spots, creating super-resolution, and impute expression levels over the entire tissue sections.

**DEEPsc** [57] uses a deep learning framework to transfer the spatial information from an ST assay onto a scRNA-seq dataset assayed from the same tissue. For each cell in the scRNA-seq data and each of the spatial spots in the ST data, a score (ranging between 0 and 1) is calculated, proportional to the probability that the cell belongs to a particular spot. To this end, a fully connected neural network is trained, which takes inputs from two vectors of equal length: one corresponding to the dimensionally reduced gene expression values of a given cell and one corresponding to the "features" of each spatial spot. The said "features" are defined to be the gene expression values of the spots in the spatial transcriptomic data, reduced to the same number of dimensions as the scRNA-seq data. The model is trained using ST assay only; then, a simulated scRNA-seq dataset is generated by adding Gaussian-distributed random noise to the ST data.

**stPlus** [58] is a reference-based autoencoder model for enhancing ST data. stPlus takes both ST data and reference scRNA-Seq data as the input. stPlus consists of three steps. First, the top 2000 highly variable genes from the scRNA-Seq dataset are selected as genes set $U$. The set of overlapping genes present in both the ST dataset and the scRNA-Seq dataset are denoted as gene set $S$. The subset of gene set $U$ in the ST data is augmented with zeros, merged with the subset of gene sets $U$ and $S$ and shuffled over cells. Second, stPlus feeds the preprocessed data into an autoencoder to learn the joint cell embeddings of ST and scRNA-Seq data. The autoencoder is trained via optimizing a two-part loss function, which consists of reconstruction loss for the subset of shared genes set $S$ in the ST data and the sparsity penalized loss of the reconstruction of the subset of genes set $U$ in the scRNA-Seq data. Finally, stPlus predicts spatial gene expression through a weighted k-NN approach based on the embeddings learned by the autoencoder. The authors showed that the predicted spatial gene expression by stPlus helped to reduce technical noise and achieved improved cell type clustering over the original ST data.

## Concluding Remarks

Many novel computational methods have been developed to tackle the challenges in computational ST. In this survey, we covered the advances in artificial intelligence for different aspects of ST analysis, including selecting SVGs, clustering analysis of spots or genes, communication analysis, cell type deconvolution, and enhancement of spatial gene expression. Of the available methods, deep learning based on neural networks are the dominant type. The flexible architecture of neural networks makes them naturally desirable candidates for building sophisticated models to analyze ST data. As the field of spatial omics continues to develop, computational ST analysis calls for more pipeline methods that can perform multiple analysis tasks and have the flexibility for integrative analysis with other data types, such as scRNA-Seq, H&E images, single cell multi-omics data [59]. Given the pace of these methods' development, a benchmarking effort is usually lacking or very limited. Thus more comprehensive comparison studies are also needed to provide researchers with valuable guidelines to choose appropriate analysis methods for various ST technologies [60].

## Competing Interests

The authors declare no competing interests.

## Acknowledgements

This work was supported by grants from the National Library of Medicine (NLM; Grant No. R01 LM012373), and the National Institute of Child Health and Human Development (NICHD; Grant No. R01 HD084633) awarded to L.X. Garmire.

# Figures and Tables

Table 1: Summary of AI methods in Spatial Transcriptomics Analysis

| Method Category | Method Name | Algorithm | Input |
| --- | --- | --- | --- |
| SVG detection | SOMDE | self-organizing-maps; Gaussian Process | ST data |
|  | scGCO | graph cut; markov random field | ST data |
| Clustering | SEDR | autoencoder; deep generative model | ST data |
|  | coSTA | convolutional neural | ST data |

|  | | network | |
| --- | --- | --- | --- |
|  | STGATE | graph attention autoencoder | ST data |
|  | RESEPT | graph autoencoder; deep convolutional neural network | ST data or RNA velocity |
|  | spaGCN | graph convolutional neural network | ST data; H&E images (optional) |
|  | stLearn | deep convolutional neural network | ST data; H&E images |
|  | spaCell | deep convolutional neural network; autoencoder | ST data; H&E images |
| Communication Analysis | GCNG | graph convolutional neural network | ST data |
|  | NCEM | deep generative model | single-cell ST data |
|  | MISTy | ensemble machine learning | single-cell ST data |
| Deconvolution | Tangram | soft mapping | ST data; sn/sc RNA-Seq; H&E images (optional) |
| | DestVI | deep generative model | ST data; scRNA-Seq |
| | CellDART | Adversarial Discriminative Domain Adaptation (ADDA) | ST data; scRNA-Seq |
| | DSTG | graph convolutional neural network | ST data; scRNA-Seq |
| Enhancement & Imputation | XFuse | deep generative model | single-cell ST data; histological images |
| | DeepSpaCE | convolutional neural network | single-cell ST data; H&E images |
| | DEEPsc | neural network | ST data; scRNA-Seq |
| | stPlus | autoencoder | ST data; scRNA-Seq |

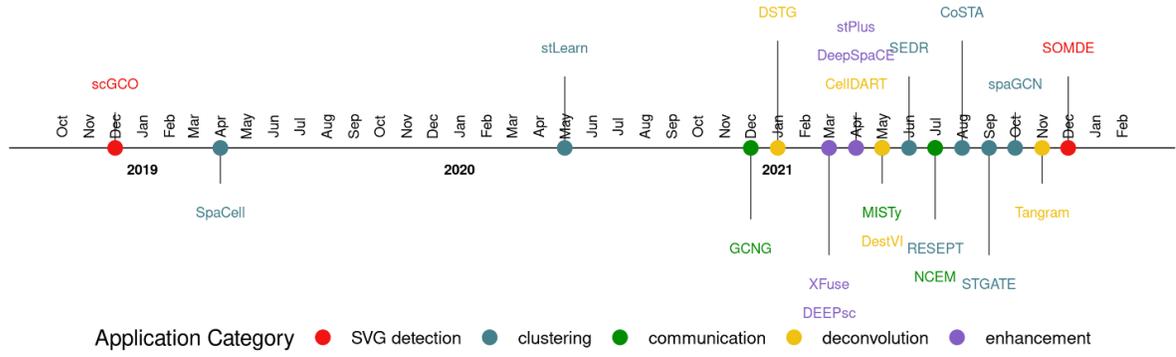
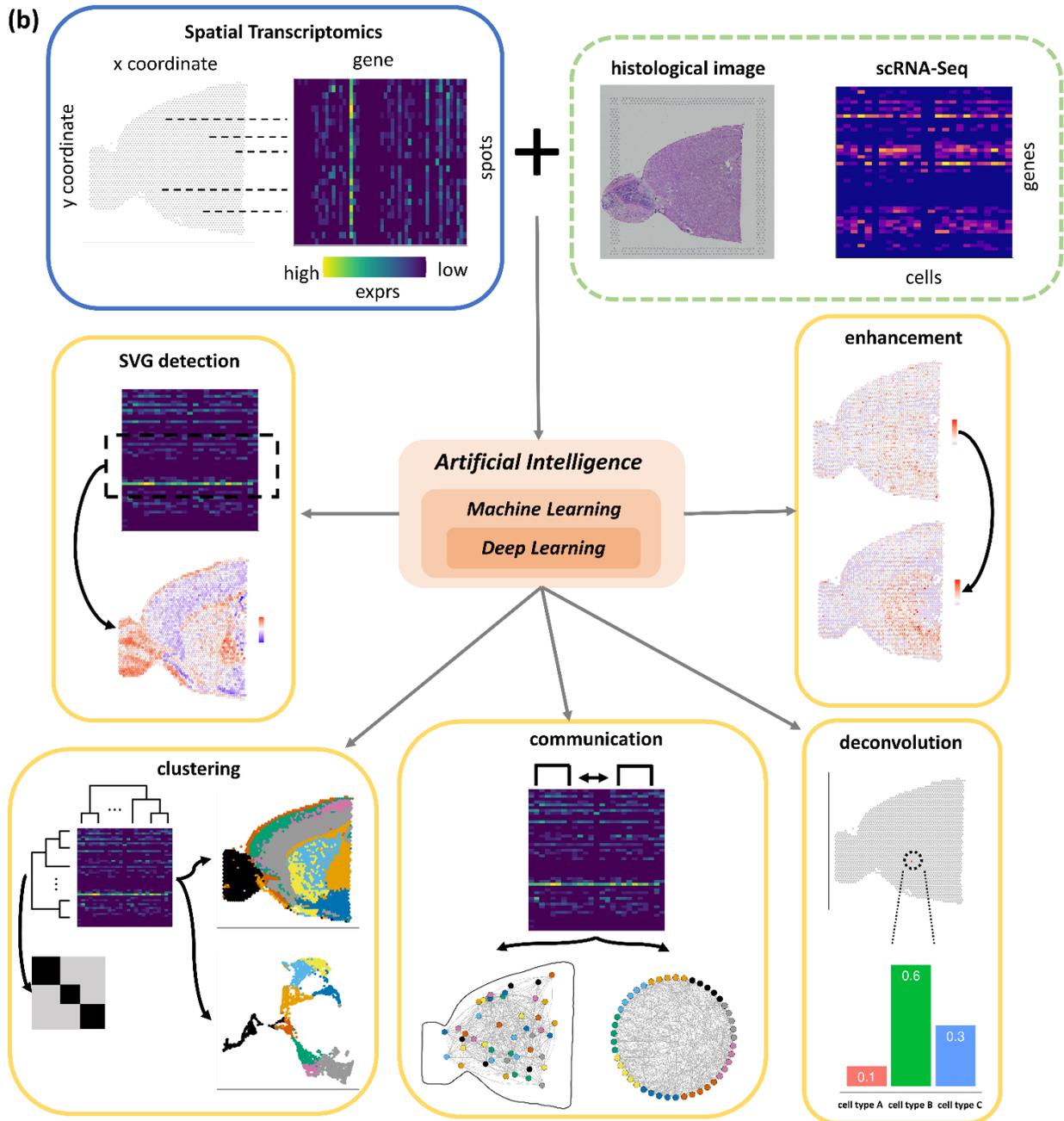

**Figure 1**. Overview of AI methodologies and application areas in ST data analysis. **(a)** Timeline of emerging AI methods in ST analysis, **(b)** characteristics of ST data, the potential reference datasets such as associated histology image and scRNA-Seq data, and the application areas in computational ST analysis: SVG detection, clustering, communication analysis, deconvolution, and enhancement.

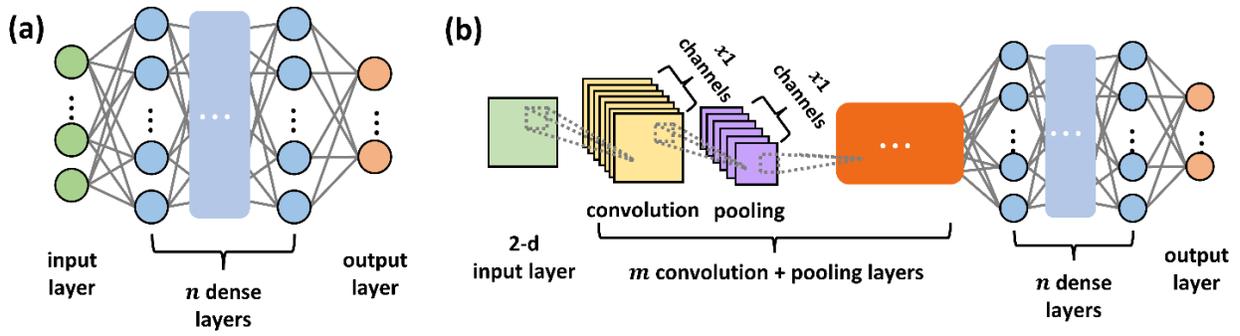

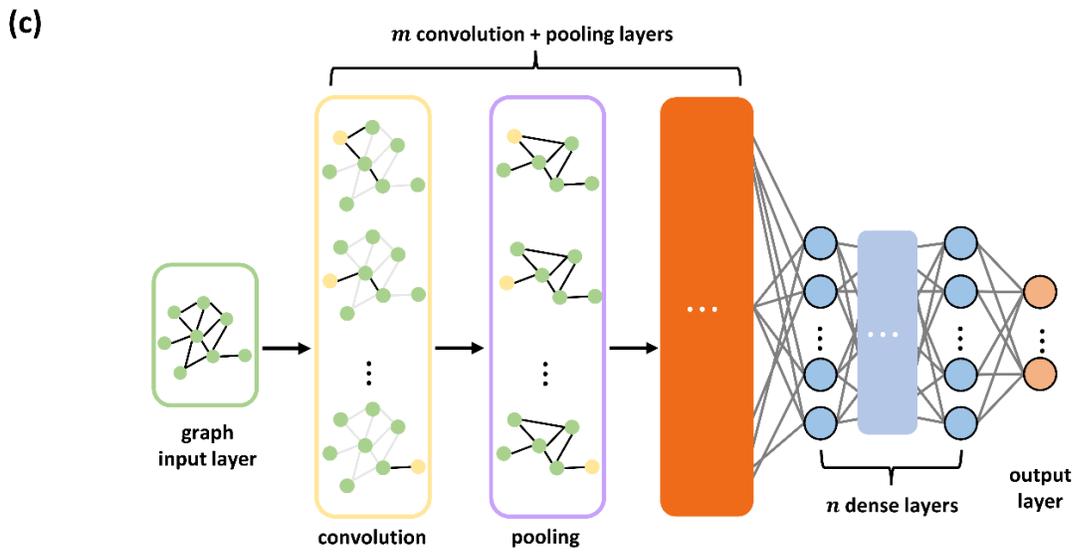

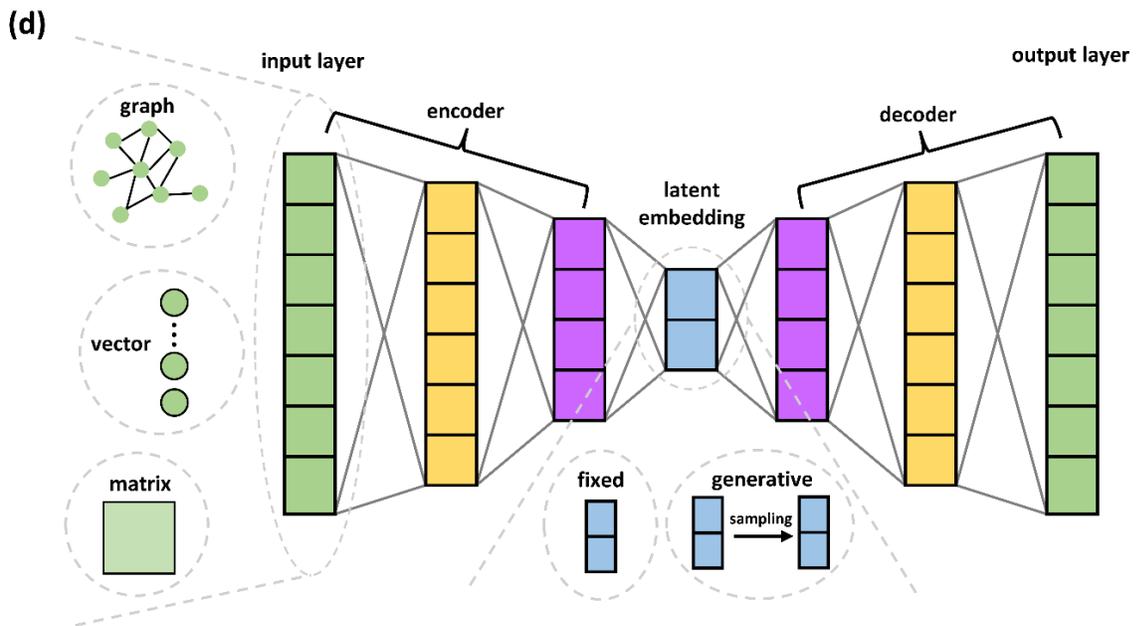

**Figure 2**. General schematic of **(a)** the fully connected neural network, **(b)** the convolutional neural network, **(c)** the graph convolutional neural network, and **(d)** the autoencoder.